\documentclass[letterpaper,10pt,conference]{IEEEtran}
\usepackage[square,numbers,sort&compress]{natbib}

\usepackage[export]{adjustbox}
\usepackage{algorithm}
\usepackage{algpseudocode}
\usepackage{amsfonts}       %
\usepackage{amssymb}
\usepackage{amsmath}
\usepackage{array}
\usepackage{booktabs}
\usepackage[font=small, labelfont=bf]{caption}
\usepackage{color, colortbl}
\definecolor{Gray}{gray}{0.95}  %
\usepackage{float}
\usepackage[T1]{fontenc}    %
\usepackage{graphicx}
\usepackage[colorlinks]{hyperref}
\usepackage[capitalise]{cleveref} %
\usepackage[utf8]{inputenc} %
\usepackage{lipsum}
\usepackage{microtype}
\usepackage{multirow}
\usepackage[numbers]{natbib}
\usepackage{nicefrac}       %
\usepackage[percent]{overpic}
\usepackage{pifont}
\usepackage{siunitx}
\usepackage{subcaption}
\usepackage{svg}
\usepackage{todonotes}
\usepackage{tikz}
\usepackage{url}            %
\usepackage{verbatim}
\usepackage{wrapfig}
\usepackage{xcolor}  %

\definecolor{mydarkblue}{rgb}{0,0.08,0.45}
\hypersetup{
linkcolor=mydarkblue,
citecolor=mydarkblue,
filecolor=mydarkblue,
urlcolor=mydarkblue}
\newcolumntype{H}{>{\setbox0=\hbox\bgroup}c<{\egroup}@{}}

\newcommand{\algoname}{HIP}

\renewcommand{\paragraph}[1]{\noindent\textbf{#1}}

\newcommand{\bo}{\mathbf{o}}
\newcommand{\bi}{\mathbf{i}}
\newcommand{\bl}{\mathbf{l}}
\newcommand{\ba}{\mathbf{a}}

\newcommand{\bx}{\mathbf{x}}

\newcommand{\bd}{\mathbf{d}}

\newcommand{\btau}{\boldsymbol\tau}

\DeclareMathOperator*{\argmin}{argmin}
\title{\LARGE\bf Hybrid Imitative Planning with \\ Geometric and Predictive Costs in Off-road Environments}

\author{
Nitish Dashora$^{*1}$, Daniel Shin$^{*1}$, Dhruv Shah$^{1}$, Henry Leopold$^{2,3}$, David Fan$^{2,4}$ \\ Ali Agha-Mohammadi$^{2}$, Nicholas Rhinehart$^{1}$, Sergey Levine$^{1}$ \\
\textit{${}^{1}$UC Berkeley, ${}^{2}$NASA Jet Propulsion Laboratory, ${}^{3}$University of Waterloo, ${}^{4}$Georgia Institute of Technology}
}

\begin{document}
\makeatletter
\let\@oldmaketitle\@maketitle%
\renewcommand{\@maketitle}{\@oldmaketitle%
    \centering
    \setcounter{figure}{0}
    \includegraphics[width=0.85\linewidth]{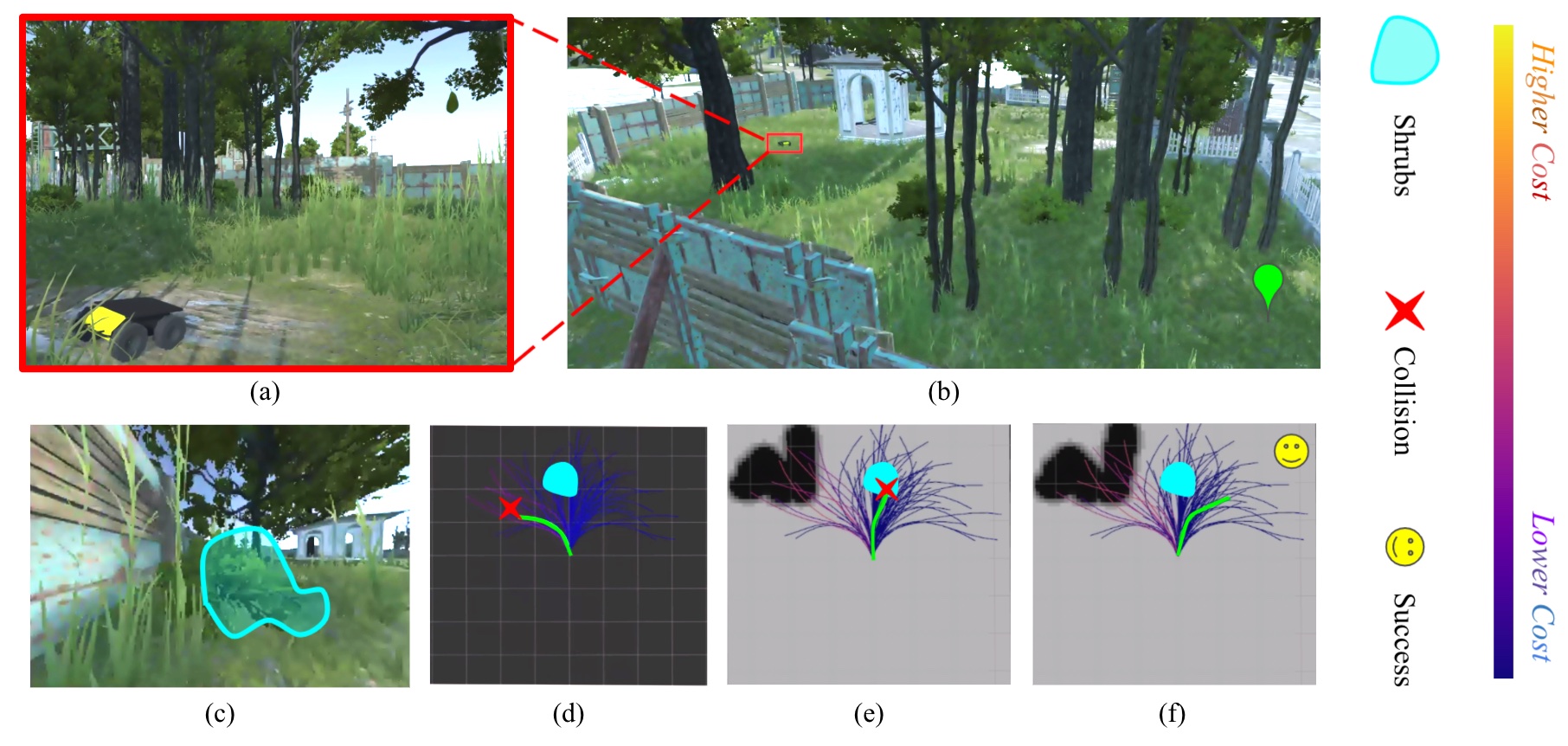}
    \vspace*{-1em}
    \captionof{figure}
    {
    Our approach combines a learning-based method and a geometry-based planner to plan obstacle-free trajectories for a Husky robot in a deployment area with obstacles (trees, grass, bushes) as seen in \textbf{(a)}. An example navigation task is shown in \textbf{(b)}, where the robot (marked in red) must navigate to the goal (marked in green) while avoiding obstacles on its path. On its path to the goal, the robot often encounters a scenes like \textbf{(c)} where there are non-traversable elements like shrubs (highlighted in cyan) and novel obstacles like a wall. 
    Learning-based methods successfully avoid previously seen obstacles but can struggle with novel obstacles, leading to a collision with the wall \textbf{(d)}. While a geometric approach can avoid this by predicting the collision, it is unable to reason about the traversability of large shrubs that are hard to identify \textbf{(e)}. Note that the cyan markers are shown for illustration only and are not available to the robot. Our approach, HIP, combines attributes from both these methods and successfully plans a collision-free trajectory \textbf{(f)}. Videos of our results are hosted at \href{sites.google.com/view/hybrid-imitative-planning/}{\small sites.google.com/view/hybrid-imitative-planning/}.
    }
    \vspace*{0.25em}
    \label{fig:overview}
}
\algrenewcommand\algorithmiccomment[2][\normalsize]{{#1\hfill\(\triangleright\) #2}}
\makeatother
\maketitle

\begin{abstract}
Geometric methods for solving open-world off-road navigation tasks, by learning occupancy and metric maps, provide good generalization but can be brittle in outdoor environments that violate their assumptions (e.g., tall grass). Learning-based methods can directly learn collision-free behavior from raw observations, but are difficult to integrate with standard geometry-based pipelines. This creates an unfortunate conflict -- either use learning and lose out on well-understood geometric navigational components, or do not use it, in favor of extensively hand-tuned geometry-based cost maps. In this work, we reject this dichotomy by designing the learning and non-learning-based components in a way such that they can be effectively combined in a self-supervised manner. Both components contribute to a planning criterion: the learned component contributes predicted traversability as rewards, while the geometric component contributes obstacle cost information. We instantiate and comparatively evaluate our system in both in-distribution and out-of-distribution environments, showing that this approach inherits complementary gains from the learned and geometric components and significantly outperforms either of them.
\end{abstract}

\section{Introduction} \label{sec:introduction}

How can we enable a robot to swiftly traverse open-world environments while minimizing heuristic and time-intensive hand-engineering, like those depicted in \cref{fig:overview}? The robot should receive coarse goal direction from a human supervisor, and use this direction, along with its sensor suite and prior experience, to make its own decisions about what actions to take to reach the destination. A solution to this problem would enable users to direct robots across unfamiliar territory without requiring them to significantly change or tune components of the system. For example, imagine a rescue worker tasking a search-and-rescue robot to quickly search a series of locations in an unmapped dense forest. A major challenge to developing such a system is enabling it to both draw on prior experience and adapt its behavior to new environments. While learning is a powerful way to deal with open-world environments, most learning-based methods studied are difficult to integrate with non-learning-based navigation pipelines. This creates an unfortunate ``either-or" dichotomy -- either use machine learning and re-design the entire navigation stack around it, or do not use it, in favor of extensively hand-tuned geometry-based cost maps and structured sensors. The key idea of our approach is to construct an algorithm by designing the learning and non-learning-based components in a way such that they can be effectively, and easily, combined and created without labeling any data. \cref{fig:overview} depicts this synthesis.
 
One classic approach is to perform online geometric mapping and traversability estimation and then use these estimates, along with a hand-tuned cost function, for planning feasible paths \citep{fan2021step, thakker2021autonomous, otsu2020fast,papadakis2013terrain}. With careful design and incorporation of prior knowledge into the decision-making pipeline, these approaches, which we term ``geometry-based'' and ``geometric costmap''-based, are a standard and performant approach. Their main drawback, however, is a general inability to automatically tune the costmap parameters with the inclusion of additional experience. For example, if the costmap assumes all densely-populated points correspond to rigid and untraversable obstacles, traversable tall grass may cause the robot to be too conservative and avoid the grass; if a height threshold is included to compensate for this effect, then small, untraversable obstacles like rocks may cause the robot to be too aggressive and get stuck on the rocks. Another approach to robot navigation is learning-based, e.g. goal-conditioned imitation learning and policy-based reinforcement learning, yet these are difficult to integrate with real-world robotic systems because they are inscrutable and directly output actions. Can we build on both lines of work by addressing their drawbacks in a principled way? 
 
\begin{figure}[h] 
\centering
\resizebox{\columnwidth}{!}{
\includegraphics[width=\textwidth]{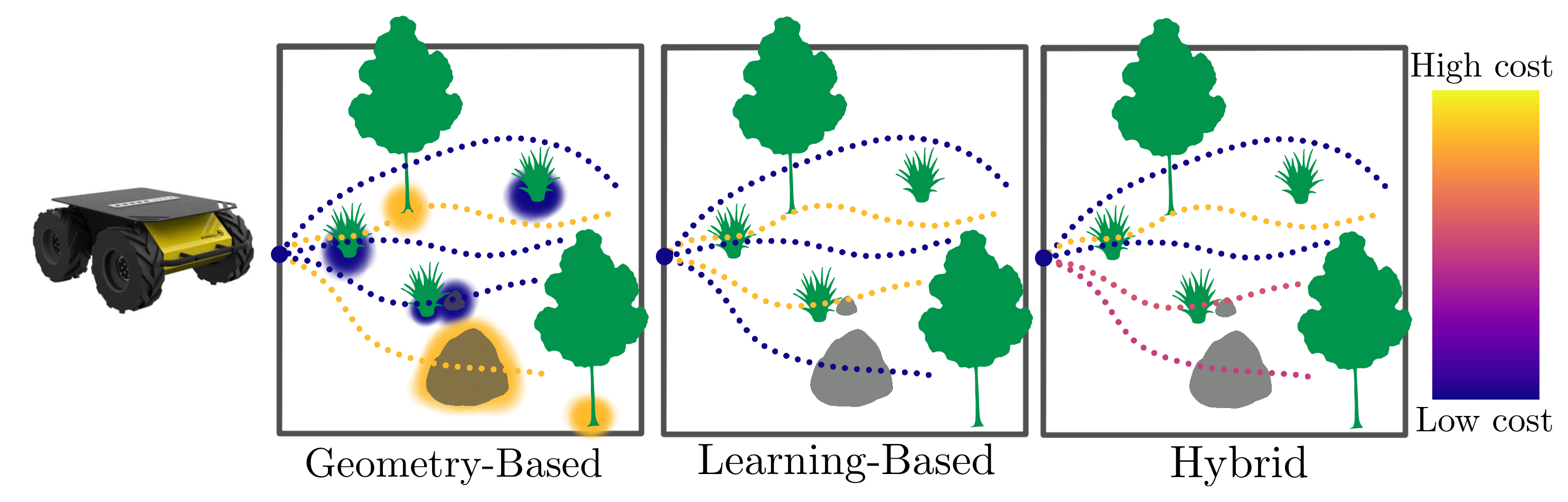}
}
\caption{\textbf{Illustration of example trajectory plan costs:} The geometric costmap approach is adept at identifying
obstacles like large trees and rocks, but it may fail to assign high costs to terrain with impediments that are only visually perceivable, since the LiDAR input often misses small obstacles.
The learning-based planner uses its prior experience to reason about the traversability of objects and terrain with which it is experienced, such as traversable grass, small impassable objects, and trees. However, it may make mistakes when it encounters drastically novel obstacles, such as painted walls or automobiles. By designing each of these components in a way that is amenable to combination, the hybrid planner can accrue benefits from each of its components. In this example, it adeptly identifies large obstacles and incorporates the data-driven experience from the learned planner to estimate that the direct plan across the small, compliant obstacles is traversable.}
 \label{fig:method_cartoon}
 \vspace*{-0.75em}
\end{figure}

The main insight in our work is that, instead of trying to reconcile conflicting \emph{actions} commanded by geometry-based and learning-based components, we can instead utilize both components to contribute terms to a shared navigational cost function. This cost function represents the hypotheses of both methods about which future trajectories will or will not lead to collisions. Once a shared cost function combining geometry-based and learning-based reasoning has been produced, a standard planning method can decide on the best path to take informed by this cost. We summarize this idea, and an intuition for why the geometry-based and learning-based components might provide complementary strengths, in \cref{fig:method_cartoon}.

The key contribution of our work is a framework to combine learned costs with a geometric costmap into a shared navigational cost function, enabling a system that can leverage the ability of learning-based methods to pick on subtle cues from high-dimensional observations and the reliability and generalization capabilities of geometric methods. We instantiate and empirically evaluate our system in a variety of in-distribution and novel environments in a high-fidelity simulator. We find that our hybrid approach is able to combine the benefits of its components and out-perform either of them in challenging domains, significantly improving performance in out-of-distribution scenes.

\section{Hybrid Imitative Planning with \\ Geometric and Predictive Costs} \label{sec:method}

In our problem setting, a mobile robot receives high-dimensional sensory observations and global localization information, and is tasked with using these sources of information to navigate to a provided goal in an unmapped, off-road environment partly populated with untraversable terrain (e.g. large and small obstacles, steep hills). Let $\bo_t=(\bi_t,\bl_t,\bx_t)$ denote the observations that are available to the robot, where $\bx_t \in \mathbb R^3$ is the robot's position estimated by odometry, $\bi_t \in \mathbb [0,1]^{H_i\times W_i\times 4}$ is an RGB-D image, and $\bl_t \in \mathbb R^{N_t \times 3}$ is a point cloud provided by a LiDAR sensor. Let $\bd \in \mathbb R^2$ denote the provided goal. Let $\btau \in \mathbb R^{2H}$ denote a trajectory of potential future positions: $\btau\doteq{\bx}_{t+1:t+H}$, and $\btau_i$ denote $\bx_{t+i}$. 

\begin{figure}[ht] 
\centering
\resizebox{.75\columnwidth}{!}{
\includegraphics[width=\textwidth]{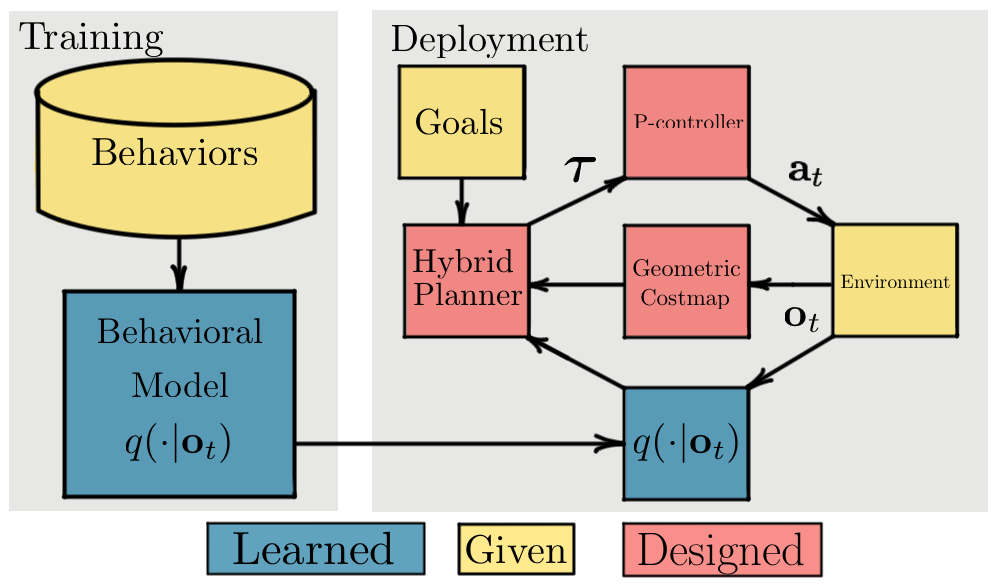}
}
\caption{\textbf{\algoname~system diagram:} A dataset of trajectory demonstrations is used to learn a model of traversability and behavior. The model is combined with a geometric costmap into the hybrid planner, which plans a trajectory towards a received goal.}
 \label{fig:flowchart}
\end{figure}

Our main idea is to design a system that allows learned models to \emph{easily integrate} with standard geometric reconstruction and planning pipelines. Learned costs and represented as a \emph{function of trajectories} and learned via density estimation to imitate collision-free data, and geometric costs are represented as a \emph{function of positions}, which can be composed into a function of trajectories, and then combined into a combined single planning criterion. Finally, destination directives are represented as positions and included as a final cost. We term our method Hybrid Imitative Planning (\algoname), and present a system diagram in \cref{fig:flowchart}. While we expect the capability to integrate these separate components to be generally useful, we have some intuition about how the utility may manifest. Geometric costmap-based planning may confer safety around large, easily-sensed objects, like boulders and trees, yet it cannot incorporate subtle visual cues, and more generally, cannot be improved from additional robot experience.  Learning-based planning may offer the capability to identify terrain directly estimated as traversable from experience, such as grass and small bushes, yet it may be risky around objects that appear significantly different than those in the training data. In designing these two approaches into a single system, we expect the system to be zero-shot robust to objects that appear significantly different from the training data and able to identify subtle aspects of the terrain that afford or inhibit traversability. In this section, we present an abstract overview of how geometric traversability costs can be integrated with learned models, and then present a specific instantiation of a navigation system based on this design in \cref{sec:instantiation}.

\paragraph{Designing geometric costmaps.} We assume the geometry-based component is available as a \emph{costmap}: $C_t(\cdot):\mathbb R^2 \mapsto \mathbb R$. This decision enables interpretable design of geometric heuristics $g(\bl_t)=C_t$ that process the LiDAR into a function of positions. Let us denote the composition of the costmap generation and evaluation of it at a position $\bx_t$ as ${C_{\text{costmap}}(\bx_t; \bo_{\leq t})\doteq g(\bl_t)(\bx_t)}$. We defer discussion of how $g$ is implemented to \cref{sec:instantiation}.

\paragraph{Designing learned costs.} In order to combine geometric and learned costs, we must construct a learner that processes the high-dimensional sensor data to assigns costs to trajectories. Our learning-based costs aim to \emph{predict whether a given trajectory would be collision-free} -- whether it corresponds to a traversable path. Formally, we construct the learning-based component as a conditional \emph{probability density function} of trajectories $\btau$: ${q(\btau | \bo_{\leq t}): \mathbb R^{2H} \mapsto \mathbb R}$, and learn $q$ from data that excludes collisions. By satisfying this assumption and training $q$ via a maximum likelihood objective, $q$ should assign high values to a subset of the \emph{traversable} trajectories and low values to this subset's complement, which will include \emph{untraversable} trajectories. Given $q$, we can construct costs as ${C_\text{learned}(\btau; \bo_{\leq t})\doteq-\log q(\btau | \bo_{\leq t})}$. Note that because $q$ is a function of trajectories, not positions, it has a higher modeling capacity than if it were parameterized by a costmap. We defer further discussion of how we learn $q$ to \cref{sec:instantiation}.

\paragraph{Designing directive costs.} In order to direct the robot to the final desired destination ($\bd$), we incorporate a cost that encourages forward progress towards the final destination. Denote this cost $C_\text{directive}(\btau; \bo_{\leq t}, \bd)$. We defer discussion of how we construct $C_\text{directive}$ to \cref{sec:instantiation}. %

\paragraph{Combining geometric costmaps and learned costs.} Given our goal of integrating learning-based and geometry-based planning, we design $\pi$ to use a receding-horizon state-space planner as shown in \cref{eqn:abstract_planning_criterion}:
\begin{align}
    &{\btau}^*\!\doteq\!\argmin_{\btau \in \mathbb R^{2H}} L(\btau; \bo_{\leq t}, \bd)\! = \Big[C_\text{directive}(\btau; \bd) +\label{eqn:abstract_planning_criterion}\\ &(1-\phi) C_\text{learned}(\btau; \bo_{\leq t}) + \phi \sum_{i=1}^{H} C_\text{costmap}(\btau_i; \bo_{\leq t}) \Big].\nonumber
\end{align}

Given a planned trajectory, $\btau$, we use it to compute $\ba_t$ as a simple position-tracking PID controller, written as ${\ba_t=f(\btau)}$.

\section{Instantiating the System} \label{sec:instantiation}
\paragraph{Implementing learned costs.} As previously discussed, our learned costs are designed to be a conditional probability density function of future trajectories, $q(\btau | \bo_{\leq t})$ fit to data that excludes collisions. We implement $q$ as a ``Deep Imitative Model'' \citep{Rhinehart2020Deep}. We adapted the open-source PyTorch implementation of imitative models released in \citet{filos2020can} and adapted the input to use RGB-D. This model uses a MobileNetV2 encoder \citep{sandler2018mobilenetv2}. During training, we follow the method of \citet{rhinehart2018r2p2} to induce an upper-bound, $\eta$, on $q$ by perturbing the training trajectories with a Gaussian. We defer further data-dependent implementation details, including a full table of hyperparameters, to \cref{sec:experiments}.

\paragraph{Implementing geometric costmaps.} We use a LiDAR sensor to produce a local point cloud, from which a terrain traversal cost is computed. We use the ``costmap\textunderscore2d'' implementation provided by ROS to compute a discrete 2D costmap for the ground plane, denote $C'_\text{costmap}$ \citep{quigley2009ros}. We apply a nonlinear transformation to $C'_\text{costmap}$ so that its maximum value corresponds to $\nicefrac{\eta}{H}$. This ensures {$\sum_{i=1}^HC_\text{costmap}(\tau_i)\leq \max_{\btau} q(\btau|\bo_{\leq t})$}, which makes tuning the $\phi$ parameter simpler, as the costs are roughly in the same range. This transformation is given by {$C_\text{costmap}((x,y))=\alpha\cdot\nicefrac{\exp{C'_\text{costmap}((x,y))}}{\sum_{x',y'}\exp{C'_\text{costmap}((x',y'))}}$}.

\paragraph{Implementing directive costs.} Similar to the region goals described by \citet{Rhinehart2020Deep},
we designed a directive cost that penalizes $\btau$ with cost $\delta$ if it does not end within a particular region a short distance away from the robot, in the direction of $\bd$. We construct this region as a rectangle with width $2\mathrm{m}$, center axis in the direction of $\bd$, offset towards $\bd$ by $3\mathrm{m}$. 

\paragraph{Fast planning.} In order to quickly solve the optimization problem in \cref{eqn:abstract_planning_criterion}, we employ a trajectory library with size with $K=200$. The future trajectories (described in \cref{sec:experiments}, were clustered using k-means, and the centroid, $\bar{\btau}^k$, of each cluster was appended to a library, $\mathcal K\!=\!\{\bar{\btau}^k\}_{k=1}^K$. During inference, we perform approximation optimization of \cref{eqn:abstract_planning_criterion} using $\argmin_{\btau \in \mathcal K}L(\btau; \bo_{\leq t}, \bd)$.

\paragraph{System and environment overview.} 
We instantiate our system on a Clearpath Husky UGV deployed in a photorealistic outdoor navigation simulator. The default sensor suite on the Husky includes a 6-DoF IMU, a GPS unit for approximate global position estimates and wheel encoders to estimate local odometry. We added a forward-facing wide field-of-view RGB-D camera and a LiDAR sensor. During data collection and evaluation, we heuristically detect collisions using IMU and odometry data. The simulator consists of a Unity backend and is tightly integrated with the ROS stack on the robot. The environment consists of an obstacle-rich, enclosed geofence where dense rigid trees, bushes, and traversable tall grass are scattered throughout. While the geometric costmap may correctly identify trees as hazardous, it can fail to disambiguate traversable grass and untraversable bushes. Furthermore, it does not reason about other terrain properties that may lead to collision, such as the terrain slope. In order for the robot to succeed, it must carefully navigate through dense obstacles over traversable terrain.

\section{Related Work} \label{sec:related_work}
\paragraph{Geometry-based goal-directed navigation.}
Existing methods for traversability estimation include 
geometry-based, appearance-based, and proprioceptive systems, as categorized by \citet{papadakis2013terrain}. Geometry-based methods build a 2.5D or 3D terrain map that is used to extract features, such as the maximum, minimum, and variance of the height and slope of the terrain \cite{lee2008cost,gestalt}. Planning algorithms for such methods can take into account the stability of the robot on the terrain \cite{hait2002algorithms, thakker2021autonomous}.  Since sensor and localization uncertainty can play a large role in the construction of environment maps, various methods exist for estimating the probability distributions of states based on the kinematic model of the vehicle and the terrain height uncertainty \cite{otsu2020fast,ghosh2018pace}.  For example, \citet{fan2021step} construct distributions of traversability costs for risk-aware planning. These approaches rely on hand-crafted models to determine risks to the robot when traversing over various terrains.  However, these models often rely on simplifying assumptions, and may not consider the compressibility or compliance of geometric features, especially with respect to vegetation \cite{mccullough2017next}. In contrast, our hybrid approach allows us to leverage geometry-based goal-directed navigation in tandem with a learned model, which enables our approach to overcome misspecifications of the geometry-based heuristics.

\paragraph{Learning-based goal-directed navigation.}
A variety of learning-based methods have studied the acquisition of goal-directed behavior, broadly classed as imitation learning (IL)~\cite{dosovitskiy2017learning, codevilla2018endtoend, ding2019goal, ghosh2018pace, shah2020ving} and reinforcement learning (RL)~\cite{kaelbling1993learning, schaul2015universal, sadeghi2016cad2rl}. Goal-conditioned IL typically requires goals to be specified during training and do not extend well beyond demonstrations. A drawback to these more general IL and RL systems is that they are difficult to interpret and incorporate into existing geometry-based goal-directed navigation pipelines. 
Planning and navigation in unstructured environments can be greatly improved by learning environment traversability using prior experience, but previous approaches to explicitly representing environment traversability require expensive human supervision or traversability heuristics~\cite{kostavelis2011supervised, garcia2018learning, howard2006towards,fan2021learning}. Recent progress in using topological graphs to implicitly reason about traversability~\cite{francis2020longrange, shah2020ving, chaplot2020nts, shah2021recon} gives a promising way to learn from prior experience but has not been demonstrated for long-range navigation in complex, unstructured environments.
In contrast, our hybrid approach employs a form of self-supervised learning-based explicit \emph{trajectory} traversability estimation in tandem with geometry-based \emph{positional} traversability estimation, which enables our approach to more robustly deal with previously-unseen complex obstacles and terrains.

\section{Experiments} \label{sec:experiments}

\begin{figure}[tb] 
\centering
\resizebox{.48\columnwidth}{!}{
\includegraphics[width=\textwidth]{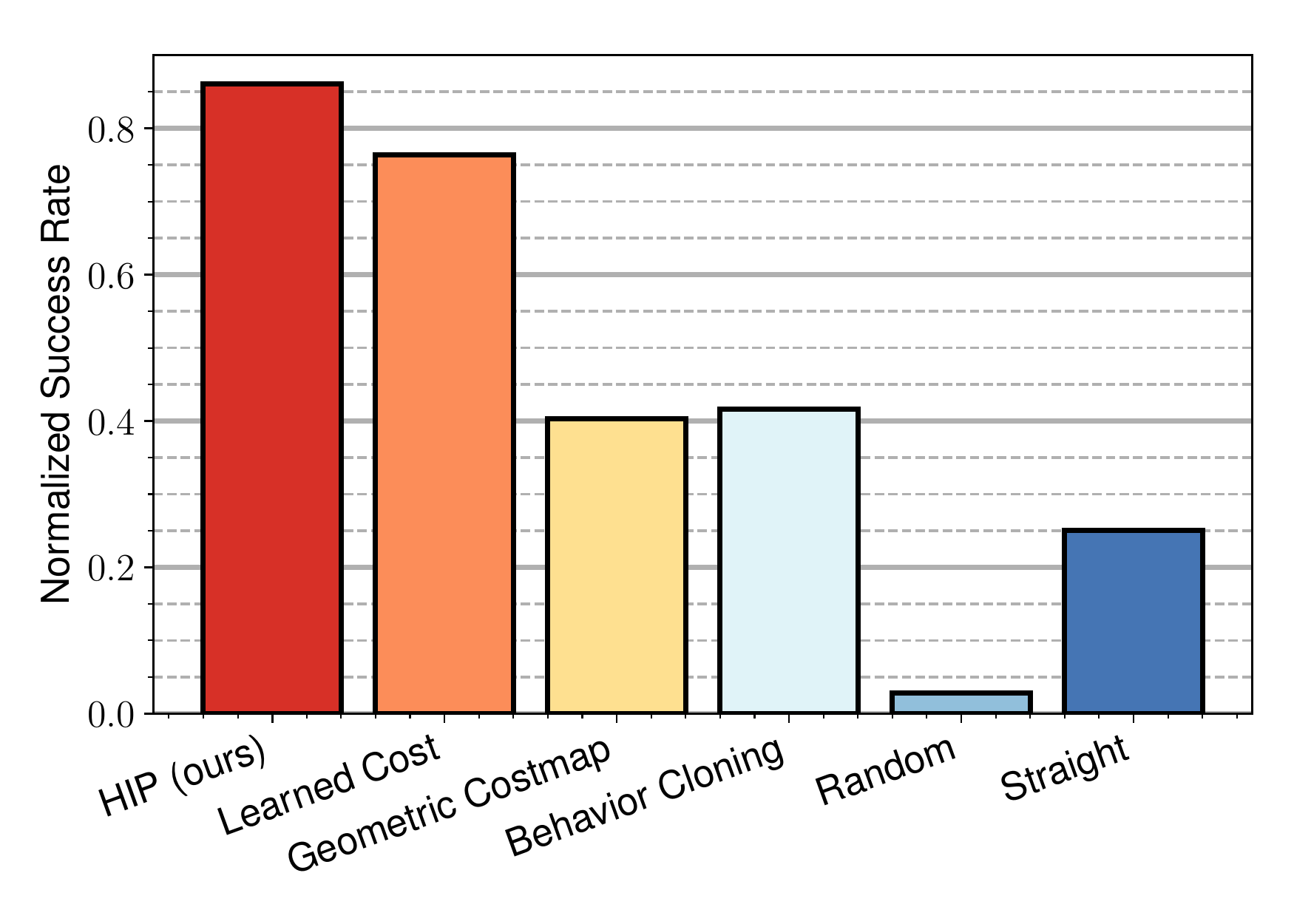}
}
\resizebox{.48\columnwidth}{!}{
\includegraphics[width=\textwidth]{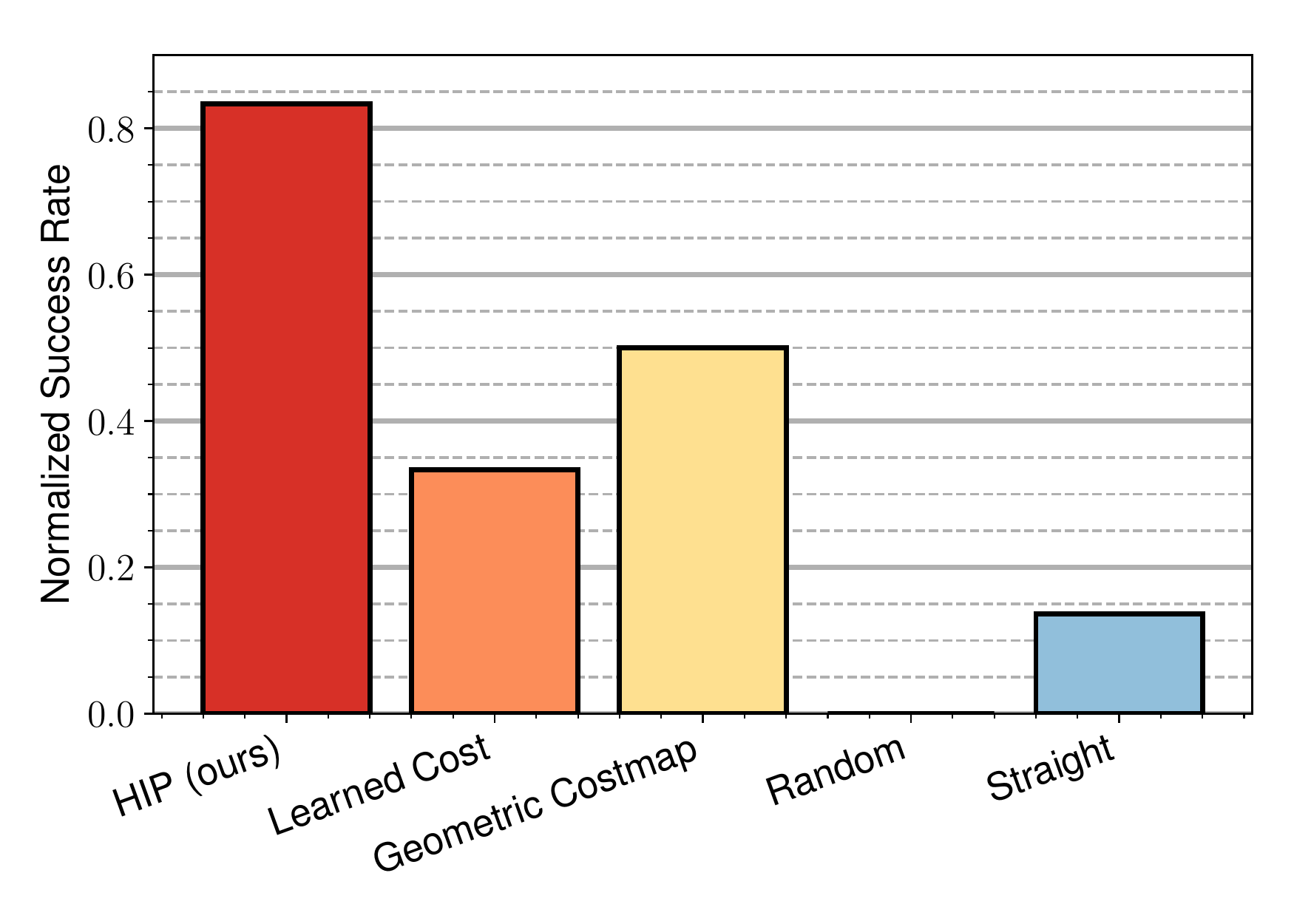}
}
\caption{\textbf{Goal-direction navigation success rates:} Our hybrid approach, \algoname, improves the task performance to the next-best approach by over 10\% in the in-distribution environment (\emph{Left}) and over 80\% in the out-of-distribution environment (\emph{Right}).}
 \label{fig:hybrid_results_barplot}
\vspace*{-2em}
\end{figure}

We designed our experiments to answer the following questions:
\textbf{Question 1:} Can our combined approach achieve collision-avoidance and navigation performance superior to its constituent components?
    \textbf{Question 2:} How does varying input modality and component weighting affect \algoname? 
    
\begin{figure*}[tb]
\centering
\includegraphics[width=0.9\textwidth]{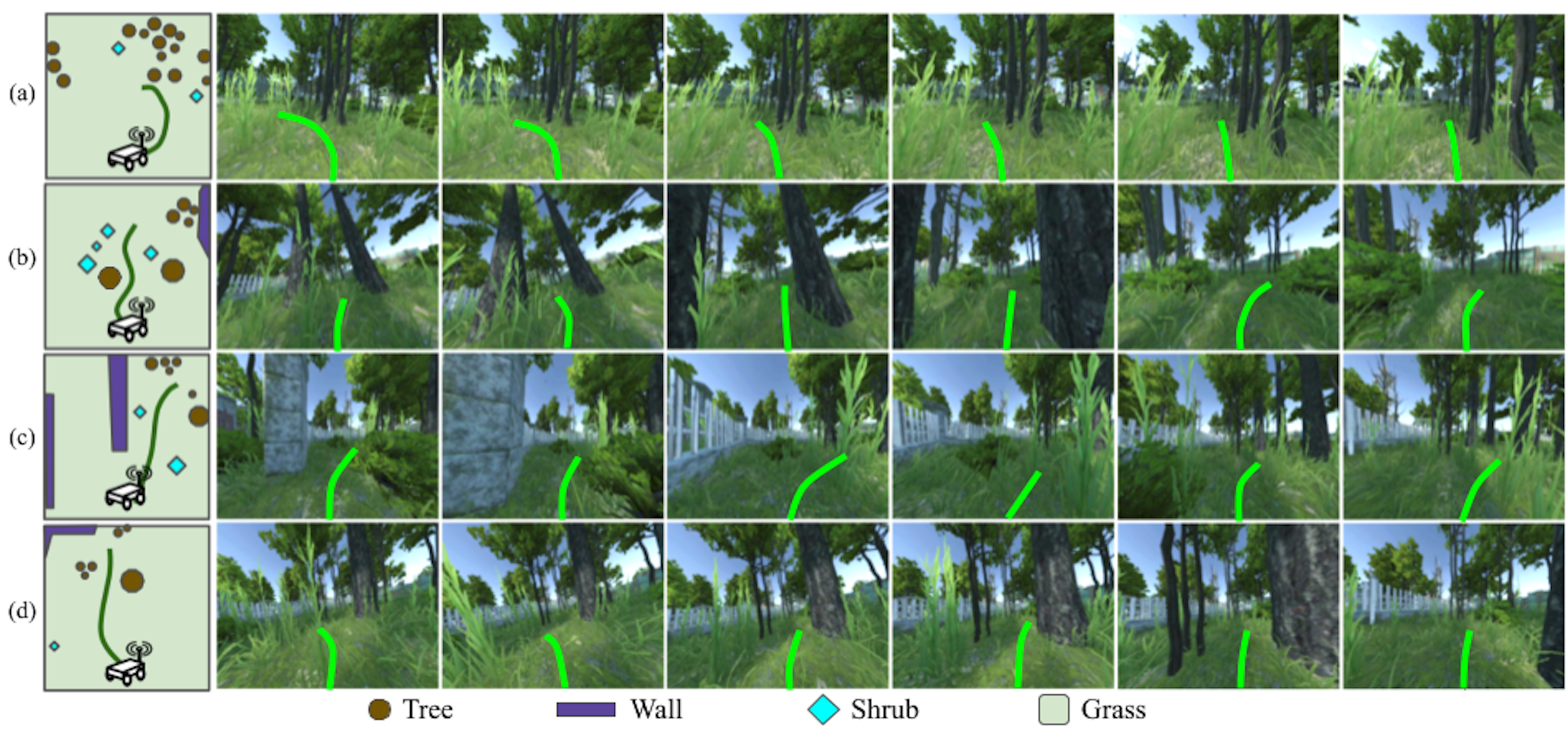}
\caption{\textbf{Example \algoname~rollouts:} We show egocentric rollouts of our method in four challenging scenes illustrated on the left (note that this view is not available to the robot). \textbf{(a)} The robot navigates through grass and avoids trees. \textbf{(b)} The robot navigates between trees and shrubs. \textbf{(c)} The robot avoids a wall and a shrub while traversing tall grass. \textbf{(d)} The robot traverses grass and between trees.}
\label{fig:filmstrip}
\vspace{-0.75em}
\end{figure*}
    
To answer these questions, we measure the rate of navigational success for randomized start-goal pairs in two different areas (in-distribution) in the simulator (which was described in \cref{sec:instantiation}). This success rate metric quantifies the performance of the model when globally directed to a determined goal; specifically, how often it succeeds in reaching the goal unharmed. We report a normalized version of this metric relative to the performance of a planner operating with a coarse-grained global obstacle map, which achieved a success rate of $0.72$. This global planner failed sometimes due to leading the robot onto hills on which it capsized and into inescapable dead-end areas with small obstacles. This ``oracle'' planner violates the problem assumption of deployment to an unmapped environment, as it has access to a map.

Towards \textbf{Question 1}, we expect the geometric costmap method to struggle with smaller untraversable obstacles, thus affecting its capability to efficiently navigate given a global direction. We expect the learner, on the other hand, to be able to generate sequences of movement through areas that are perceptually similar to traversable scenes in the training data. Since any learned model is susceptible to errors under distributional shift, we might expect the learner to sometimes fail to produce collision-free trajectories when it encounters obstacles that are visually very different from those seen in the training data.
Finally, we expect our proposed hybrid approach to harness the strengths of its components to outperform them with minimal hand-engineering effort. We expect the improvement to result primarily from superior obstacle-avoidance behavior.

Towards \textbf{Question 2}, we first investigate the effect of $\phi$, which controls the importance of the component costs on the planning criterion (\cref{eqn:abstract_planning_criterion}). By varying $\phi$ to identify the optimal value, $\phi^*$, we determine if they complement each other $(\phi^* \in (0, 1))$, or whether one dominates $(\phi^* \in \{0, 1\})$. Furthermore, through the ablation, we can understand the importance of each sensor; we expect the learner's performance to be maximized when all sensor modalities are included.

\paragraph{Baselines.} Beyond the components of our method, we include three other baselines to further contextualize our system's performance. \textbf{Behavior Cloning (BC) baseline:} We trained a goal-conditioned BC baseline using the same data the learner used, as well as a nearly identical neural-network architecture to that of the learner. The BC baseline is trained to directly predict a control given a provided goal, rather than a probability density function over trajectories. \textbf{Straight baseline:} In order to contextualize the importance of reactivity to perceptual cues, we include a baseline that plans an action to track the straight-line segment from the robot to the goal. This baseline does not process the perceptual data, and therefore cannot react to obstacles. \textbf{Random baseline:} Finally, we include a baseline that uniformly samples a random action from the robot's discretized action space, which further contextualizes the difficulty of the problem.

\paragraph{Data collection and model learning.} 
We performed the following steps to automatically collect training data in the simulator. First, we created a geofence and a generated a set of random starting poses within it. Next, we randomized the robot's starting pose among this set and drove it with ``sticky'' (executed for multiple frames) random actions.  As per our method's data requirements described in \cref{sec:method}, we automatically removed sequences that resulted in collisions. Although the explicit collision data could be employed to refine the model, we found this method to be effective even when the collision data was discarded entirely. Collisions were heuristically identified in three different ways, and the robot was randomly respawned after any collision. The first heuristic is a \textit{stuck collision}, which triggers when the robot is stationary for over 4 seconds. The second is a \textit{trapped collision}, which triggers when the robot does not move over 3m over a 10s period, which prevents tight circular movement or very slow motion. The final heuristic is a \textit{capsized collision}, which uses the IMU to identify when the robot is capsized.
As part of each recorded positional trajectory, RGB-D and odometry data was recorded. To create the set of examples for training, the data was postprocessed by subsampling at 5Hz. Each example uses $10$ past timesteps and a one RGB-D visual input, and is trained to predict $H=10$ timesteps of future positions via a maximum likelihood objective. Over $>200k$ examples were collected, with which our Learner was trained to maximize the likelihood of the ground truth odometry positions conditioned on the past odometry and visual information: $\max_\theta \mathbb E_{(\bo_{\leq t}, \tau)}\log q_\theta(\btau|\bo_{\leq t})$. \cref{tab:hparams} summarizes the hyperparameters of our method.

\begin{figure*}[h]
\centering
\includegraphics[width=0.9\textwidth]{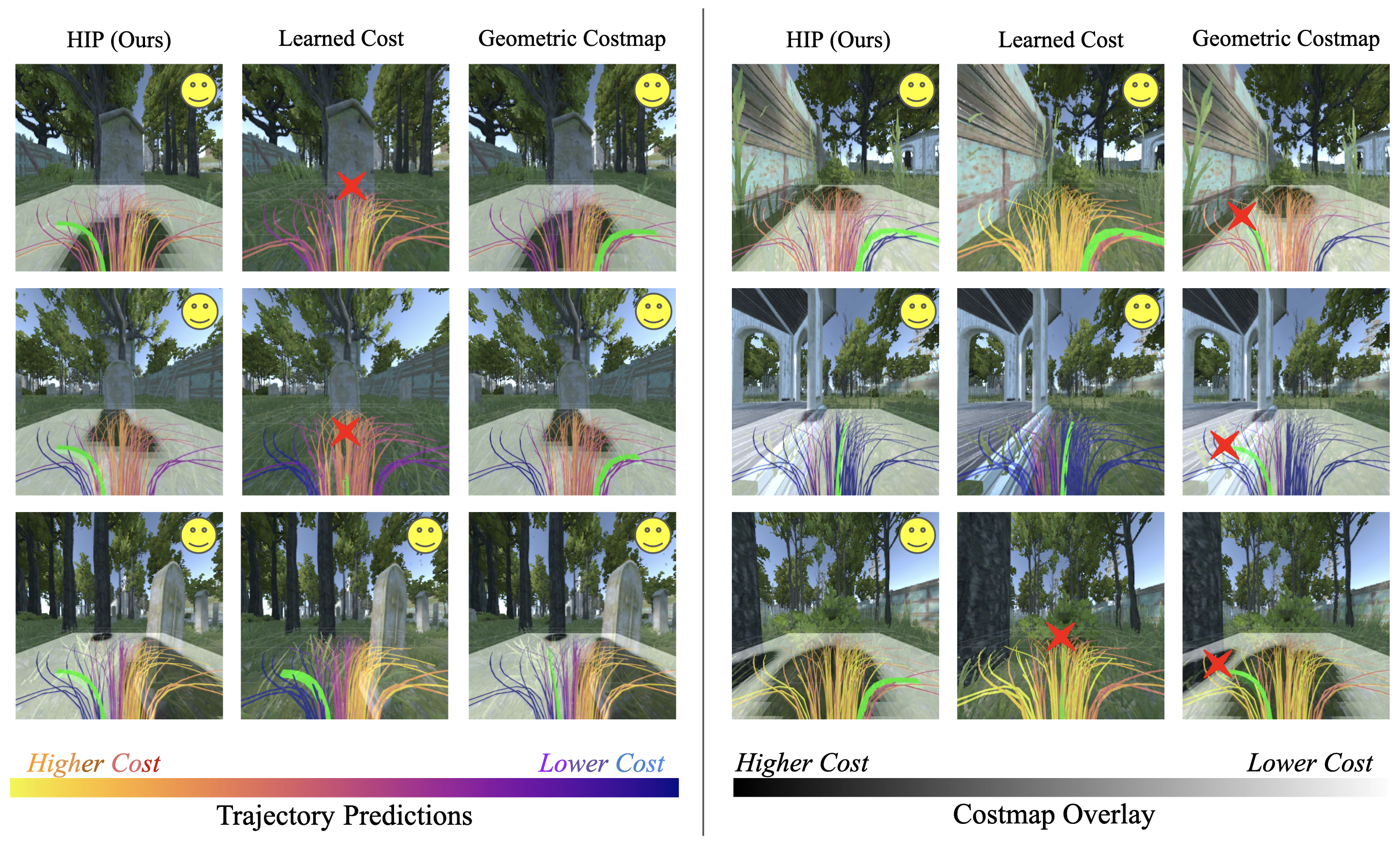}
\caption{\textbf{Qualitative results on six example scenes:} We visualize the trajectory library for the different methods and highlight the plans from each component in green. Methods which utilize a geometric costmap have an overlaid map where black represents high cost and gray represents low cost as shown in the right color map. All trajectories are color coded by their geometric cost as shown in the left color map. The left three scenes are out-of-distribution while the right three scenes are in-distribution. 
Since the geometric costmap cannot forecast the future, it may sometimes lead into obstacles just outside the planning horizon. When out of distribution, the learned planner fails to recognize unseen obstacles and can cause collision. Our hybrid planner combines the benefits of each component to navigate successfully.}
\label{fig:dcist_results}
\vspace*{-0.75em}
\end{figure*}

\begin{table}
\centering
\resizebox{0.9\columnwidth}{!}{
\begin{tabular}{lHc}
\toprule
Method & $\textit{\textbf{t}}_{avg}^{ivntn}$ & Normalized Success Rate \\   
\midrule
Learned Cost-only, RGB-D input & 41s & 0.76 \\ %
Learned Cost-only, RGB input & & 0.60 \\ %
Learned Cost-only, Depth input& & 0.65 \\ %
\bottomrule
\end{tabular}
}
\caption{\textbf{Ablation results:} We observe that both RGB and Depth information are helpful to the Learner.} \label{table:sim_ablation}
\end{table}
\begin{table}[tb]
\centering
{
\resizebox{0.95\columnwidth}{!}{
\begin{tabular}{clr}
\toprule
Hyperparam. & Value & Meaning \\ \midrule
$f_\text{Sim}$ & $30\mathrm{Hz}$ & Simulator framerate \\
$|\mathcal D|$ & $\sim200\mathrm{k}$ & Dataset size \\
$f_\tau$ & $5\mathrm{Hz}$ &  Trajectory framerate \\
$H$ & $10$  & Learner's prediction horizon \\
$H_\text{past}$ & $10$ & Learner's input horizon \\
$H_i \times W_i$ & $100\times100$ & Cropped RGB-D image dimensionality \\ 
$B$ & 32 & Minibatch size \\
$\epsilon$ & 0.001 & Learning rate \\
$\sigma$ & $0.01$ & Scale of the training perturbation \\
$\eta$ & $64$ & $\sigma$-induced upper-bound of $q(\btau|\bo_{\leq t})$ \\
$\alpha$ & $6.4$ & Costmap scaling parameter \\
$\phi^*$ & $0.75$ & Final planner cost balance \\ 
$K$ & $200$ & Trajectory library size \\
$f_L$ & $1\mathrm{Hz}$ & Replanning frequency \\
\bottomrule 
\end{tabular}
}}
\caption{Hyperparameters used in our experiments.\label{tab:hparams}}
\vspace*{-1.75em}
\end{table}

\begin{figure}[tb] 
\centering
\resizebox{.8\columnwidth}{!}{
\includegraphics[width=\textwidth]{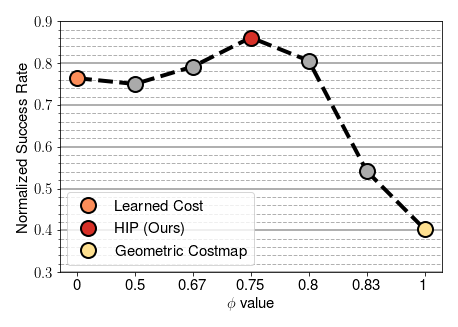}
}
\caption{\textbf{Success rate on varying $\phi$:} When no cost is incorporated ($\phi=0$), our method purely relies on learning; when the learned component is removed ($\phi=1$), our method purely relies on the costmap. Empirically analyzing a range of $\phi$ suggests that combining the components with $\phi=0.75$ results in the best performing system.}
 \label{fig:costmap_param_sweep}
\vspace*{-1.75em}
\end{figure}

\paragraph{Experimental setup.} %
The simulator we employed in our experiments was described in \cref{sec:instantiation}. To measure success rate, $\textit{\textbf{s}}_{rate}^{goal}$, the robot was randomly spawned $300$ times in the geofence with a goal at least $10\mathrm{m}$ away. A constant seed was utilized for generating start and goal points across each method. We conducted this experiment in two environments: (1) the environment from which data was collected, and (2) an out-of-distribution environment containing novel obstacles.

\paragraph{Results analysis.} %
Our primary results are shown in \cref{fig:hybrid_results_barplot}. The na\"{\i}ve baselines -- Random and Straight -- illustrate the difficulty of the problem. We find our method to be the most performant. The Learner significantly outperforms the Behavioral Cloning baseline, as it represents multiple possible futures and defers goal-conditioning until test time, when the planner can select the most appropriate trajectory for the given goal. The Geometric Costmap outperforms the na\"{\i}ve baselines, and performs similarly to BC. It cannot independently model traversability for rigid objects shorter than grass, such as bushes or rocks, but is adept at identifying areas with larger, more distinguishable untraversable objects. \cref{table:sim_ablation} contains the result of the ablation analysis. The RGB-D Learner outperformed both RGB and Depth alone, thus illustrating the significance of both modalities in determining traversability. This gap is further widened in out-of-distribution environments, where the learned costmap suffers due to unreliable predictions.

In \cref{fig:costmap_param_sweep}, the results of the $\phi$ hyperparameter sweep are presented. These results affirm the efficacy of both components of our method, show that multiple $\phi$ values are performant. In \cref{fig:filmstrip}, we show example rollouts from our method navigating dense obstacles; we observed that it is capable of winding through complex areas with many obstacles if an open path exists. In \cref{fig:dcist_results}, a set of qualitative examples for our method, the Learner, and the Costmap are depicted. In these examples, we often observe the Costmap to identify a subset of the impassable obstacles, while the Learner refines the remaining viable paths to account for previous odometer positions and objects perceived in the RGB-D image (which may be undetected by LiDAR). This results in the most effective navigation around visually perceivable obstacles.

\section{Discussion} \label{sec:discussion}

We proposed \algoname, a method designed to flexibly incorporate learning-based and geometry-based components into a single cost function for goal-directed navigation in open-world off-road environments. We evaluate \algoname~in a high-fidelity simulator, and find that \algoname~shows significant improvement over both of its constituent components, as well as baselines. Ablations of sensory inputs confirm the efficacy of both RGB and depth data to the system. 
A hyperparameter sweep of the primary parameter illustrates the existence of multiple performant values.
The effectiveness of our approach illustrates that both learning-based and geometry-based components for autonomous navigation can be integrated effectively if they contribute distinct but complementary terms to a shared cost function. That is, instead of sharing \emph{control} between different types of methods, we simply add their contributions to a cost function used by a standard model-predictive control method. This approach suggests promising directions for future work, integrating other types of learned models, as well as additional sensory modalities. Furthermore, since our learned costs can be integrated with arbitrary goal representations into a standard planner, a promising direction is to further study other types of objectives that can be accomplished with our method.

\bibliographystyle{IEEEtranN}
\bibliography{references}

\end{document}